\documentclass[11pt]{article}

\usepackage{nodalida2019}
\usepackage{times}

\usepackage{latexsym}

\usepackage{multirow}

\usepackage{amsmath}

\usepackage{adjustbox}

\usepackage{flushend}

\usepackage{float}

\aclfinalcopy 

\usepackage{hyperref}
\hypersetup{breaklinks}
\hypersetup{raiselinks=false}
\hypersetup{hidelinks=true}

\title{May I Check Again? — A simple but efficient way to generate and use contextual dictionaries for Named Entity Recognition. Application to French Legal Texts.}

\author{Valentin Barriere \\
	Cour de Cassation \\
	Palais de Justice \\
	5 quai de l'horloge\\ 75001 PARIS \\
	{\tt valentin.barriere@justice.fr} \\
	\And
	Amaury Fouret \\
	Cour de Cassation \\
	Palais de Justice \\
	5 quai de l'horloge\\ 75001 PARIS \\
	{\tt amaury.fouret@justice.fr} \\
}

\date{}

\begin{document}
	\maketitle
	\begin{abstract}
		In this paper we present a new method to learn a model robust to typos for a Named Entity Recognition task. 
		Our improvement over existing methods helps the model to take into account the context of the sentence inside a court decision in order to recognize an entity with a typo. 
		We used state-of-the-art models and enriched the 
		last layer of the neural network with high-level information 
		linked with 
		the potential of the word to be a certain type of entity. 
		More precisely, we utilized the similarities between the word and the potential entity candidates in the tagged sentence context. 
		%
		The experiments on a dataset of French court decisions show a reduction of the relative F1-score error of 32\%, upgrading the score obtained with the most competitive fine-tuned state-of-the-art system from 94.85\% to 96.52\%.  
		%
		
		
		
	\end{abstract}
	
	\section{Introduction}
	
	Automatic Named Entity Recognition (NER) is a task that has been tackled and tackled over the years, because of the multitude of possible applications that flow from it. It can be useful for entity information extraction \cite{Ferre2018}, for the creation of Knowledge Bases like DBPedia or for purposes of pseudonymisation (identification and replacement) in 
	sensitive documents from the medical or the legal domain \cite{Neamatullah2008}. 
	
	In our application, the French Courts of Justice release 3800k court decisions each year. 
	The size of this number makes the manual de-identification of each court decision helpless. Hence it is mandatory to use natural language processing NER tools 
	to automatize the operation. 
	
	The domain of NER has considerably evolved in the last several years. 
	The NER models can be rule-based systems using expert knowledge \cite{Neamatullah2008}, hybrid models using linguistics and domain specific cues as features of a learning method \cite{Sutton2011, Bodnari2013} or end-to-end deep learning models using distributed learned representations of words and characters \cite{Peters2018a, Lample2016}. 
	
	Each method has its own advantages and drawbacks. The rule-based ones allow high precision but are nonetheless domain-specific, nor robust to noisy data and costly to design
	. Hybrid methods combine the robustness and the high accuracy of Machine Learning algorithms with the fine-grained information of external dictionaries or linguistic rules \cite{Cohen2004, Barriere2017b}. The deep learning approaches that achieve high performances relying on a big amount of training data are the most efficient nowadays \cite{Devlin2018}. 
	
	Nevertheless, even the most efficient systems struggle to manage with some kind of noise: the typos and misspelling \cite{Kukich1992} are common in real-world tasks, up to 15\% of the search queries \cite{Cucerzan2004}, and lower the performances of the NER systems \cite{Lai2015}. 
	
		
	
	
	
	In this paper, we propose a new method called MICA (May I Check Again) that improves the performances of a state-of-the-art NER model using contextual information by automatically generating contextual dictionaries of entities. 
	We use a two-learning-step method that: learns a first NER neural network model, then uses the first network to create a list of potential entities that will be used to create new features for each word in the last layer of the second NER neural network model. 
	We chose last layer since those new features contain high-level information regarding our task and the level of complexity increases with the depth of neural network \citep{Sanh2017}. Nevertheless, this method can also be used with a simple NER system like Conditional Random Fields, and it shows interesting results although not state-of-the-art. 
	
	The use of language-specific knowledge-source, dictionaries or gazetteers is very common for this type of task. \citet{Neelakantan2015} also proposed to learn dictionaries of entities for NER and we distinguish our work from theirs by several points. Our method does not aim to create a dictionary of entities but instead use the entities detected in the context of each sentence in order to enhance the NER model. 
	
	Finally, we also worked on the language model embeddings in order to adapt the language models and embeddings from general domain to the legal domain. For that, we refined the BiLM Flair embeddings \cite{Akbik2018} and trained the Fastext embeddings \cite{Grave2018} on a dataset of 660,000 court decisions in order to adapt the language models and embeddings from general domain to the legal domain. 
	
	
%

	
	
%
	
	
	\section{Sequence Labeling models}
	
	The method we are presenting : \vspace*{-.2cm}
	\begin{enumerate}
		\item Learn a Vanilla model for NER \vspace*{-.2cm}
		\item For each sentence, create a list of potential entities using a context window \vspace*{-.2cm}
		\item Create a vector of similarity values for each word between the word and each type of entities \vspace*{-.2cm}
		\item Use this vector as a new feature in the last layer of the new NER neural network 
	\end{enumerate} 

	As Vanilla sequence tagger model, we chose to use the work of \citet{Akbik2018} which obtained state-of-the-art results for NER, part-of-speech-tagging and chunking. 
	This method is not specific to the use of deep learning, though adding high-level information on the last layer is perfectly adapted to our problem, but can apply to any sequence tagger model. 
	
	In order to verify this hypothesis, we also used MICA with a basic NER model composed of a Conditional Random Fields using hand-crafted features as input \cite{Peng2014}. 
	
		
	
	\subsection{Vanilla Model}
	
		
	The Vanilla model consists of a Bidirectional Long Short Term Memory coupled with a Conditional Random Field output layer (BLSTM-CRF) \cite{Huang2015a} at the word level. The input of the BLSTM-CRF is a global vector composed of the concatenation of three different embedding vectors (see Equation \ref{eq:stacking}).
	%
	
	\paragraph*{Vector Stacking} This global vector counts a contextualized word embedding vector obtained with a Bidirectional character-level Language Model and a word embedding vector learned independently of the NER task, and a character-level word embeddings learned jointly with the NER task, as shown below: 
	
  \begin{align}
		\textbf{w}_i &= \begin{bmatrix}
		\textbf{w}_i^{FastText} \\
		\textbf{w}_i^{CharBiLM} \\
		\textbf{w}_i^{Char} \\
		\end{bmatrix}
		\label{eq:stacking}
  \end{align}
	where $\textbf{w}_i^{CharBiLM}$ is the precomputed Bidirectional character-level Language Model from \citet{Akbik2018}, $\textbf{w}_i^{FastText}$ is the precomputed FastText from \citet{Grave2018} and $\textbf{w}_i^{Char}$ the character-level word embedding learned during the task \cite{Ma2016}. 
	
	\paragraph*{BLSTM-CRF} 
	
	For each word, we'll obtain as output of the BLSTM a vector $\textbf{r}_i$ (see Equation \ref{eq:output_state} where $\textbf{r}_i^f$ and $\textbf{r}_i^b$ are respectively the forward and backward output states). 
	
	\begin{align}
		\textbf{r}_i &= \begin{bmatrix}
		\textbf{r}_i^f\\
		\textbf{r}_i^b \\
		\end{bmatrix}
		\label{eq:output_state}
	\end{align}
	
	The sequence of $\textbf{r}_i$ vectors is used as observations for the CRF \cite{Lafferty2001}.  
	
	More details on the model can be found in \cite{Akbik2018}. 
	

%
%

	\subsection{MICA} \label{sect:typo_robust}
	
	\paragraph*{Vanilla model} Once the training of the Vanilla NER model over, we create a new model with the same architecture and initialize its weights with the ones of the trained Vanilla except for the CRF last layer. 
	
	\paragraph*{Similarity Vector} For every sentence $sent$ the new model sees, the Vanilla model sees a bunch of sentences from its neighborhood and create a dictionary of local entity candidates $D_{sent}$ (see Equation \ref{eq:dico}).
	
	\begin{align}
		D_{sent} &= \begin{bmatrix}
		\text{PER} : [c_{\text{PER}}^1, ..., c_{\text{PER}}^{L_{\text{PER}}} ] \\
		\text{PRO} : [c_{\text{PRO}}^1, ..., c_{\text{PRO}}^{L_{\text{PRO}}} ] \\
		\text{LOC} : [c_{\text{LOC}}^1, ..., c_{\text{LOC}}^{L_{\text{LOC}}} ] \\
		\text{DATE} : [c_{\text{DATE}}^1, ..., c_{\text{DATE}}^{L_{\text{DATE}}} ] \\
		\end{bmatrix}
		\label{eq:dico}
	\end{align}
	
	We can create for each word $w_i \in sent$ a vector $\textbf{s}_i$ containing the potentiality of a word being an entity of each type, computing similarity with the Damerau-Levenshtein distance $d_L$ \cite{Damerau1964, Levenshtein1966}, and the longest common string $LCS$. The Damerau-Levenshtein distance is a derivative of the Levenshtein one known to be useful for misspellings detection. For each entity type, we compute the Damerau-Levenshtein similarity between the word $w_i$ and the entity candidates, and take the maximum value. 
	We also used a similarity based on the longest common string between the word and the most similar entity candidate $c_{\text{ENT}}^*$.
	
	
		\hspace*{-1.6cm}
	\begin{align}
		\textbf{s}_i &= \begin{bmatrix}
		\underset{l}{\max}(\text{Lev}(w_i, c_{\text{PER}}^l)) +\text{ LCS}(w_i, c_{\text{PER}}^*) \\
		\underset{l}{\max}(\text{Lev}(w_i, c_{\text{PRO}}^l)) +\text{ LCS}(w_i, c_{\text{PRO}}^*) \\
		\underset{l}{\max}(\text{Lev}(w_i, c_{\text{LOC}}^l)) +\text{ LCS}(w_i, c_{\text{LOC}}^*) \\
		\underset{l}{\max}(\text{Lev}(w_i, c_{\text{DATE}}^l)) +\text{ LCS}(w_i, c_{\text{DATE}}^*) \\
		\end{bmatrix}
		\label{eq:vector_sim}
	\end{align}

	
	\paragraph*{Enriched CRF} Then we stack the vector $s_i$ to the previous $r_i$ vector which is the input of the CRF (see Equation \ref{eq:vector_added}). 
	
	\begin{align}
		\textbf{r}_i^{\text{enhanced}} &= \begin{bmatrix}
		\textbf{r}_i^f\\
		\textbf{r}_i^b \\
		\textbf{s}_i \\
		\end{bmatrix}
		\label{eq:vector_added}
	\end{align}
	

	\subsection{Simple CRF} \label{sect:crfconll}
	
	The MICA method does not necessarily need to be used with a neural network although it is appropriate, so we also experimented MICA with a simple NER model. We tested it using a simple baseline model: a Conditional Random Fields using classical hand-crafted features as input. We used all the features of the CONLL 2002 NER Tutorial of \citet{Peng2014} except the parts of speech that are not given in our dataset. 
	
	The configuration stays the same, with the handcrafted features vector $\textbf{r}_i$ concatenated with the similarity vector $\textbf{s}_i $ as input of the CRF. 
	

	\section{Experiments} \vspace*{-.2cm}
	
	We tested three kind of models, that were all build upon a state-of-the-art performing system for Named Entity Recognition \cite{Akbik2018} that we call Vanilla for reasons of simplicity. The Vanilla model is a BiLSTM-CRF taking as input different kinds of embeddings learned on general text data. 
	We compared the Vanilla model with a model using embeddings that were fine-tuned or learned on legal text from the same domain that the text in our NER dataset. 
	Eventually, we compare those baseline models to our models with the CRF layer enhanced with high-level similarity information. 
	
	All the models were compared on the same dataset, with the same split between the train, validation and test datasets. Each set constitutes respectively approximately 80\%, 10\% and 10\% of the full dataset. 
	
	All the models have been implemented using Pytorch \cite{Paszke2017} and based on the Flair toolbox \cite{Akbik2018}. The simple CRF had been implemented using pycrfsuite \cite{Peng2014}.

	\subsection{Dataset}
	
	Our dataset is composed of 94 of real court decisions for a total of 11,209 sentences and 276,705 tokens. It has been manually annotated by a unique law expert regarding the following 4 types of entities: 
	\begin{enumerate}
		\item \textbf{PER}: first and last name of the individuals, \vspace*{-.2cm}
		\item \textbf{PRO}: first and last name of the court members and attorneys, \vspace*{-.2cm}
		\item \textbf{LOC}: addresses concerning birthplaces and residences,  \vspace*{-.2cm}
		\item \textbf{DATE}: dates of birth. \vspace*{-.2cm}
	\end{enumerate}

	Following the protocol of CONLL-2003 \cite{Tjong2003}, the dataset has been annotated with the BIO scheme \cite{Ramshaw1995} and separated into a train, a development and a test dataset. The statistics of the subdivided sets are shown in Table \ref{tab:dataset}. 
	
	Examples of decision after anonymization of the PER, LOC and DATE classes can be found on the Internet website of 
	
	Due to the facts that a second annotation pass would be costly and that the court decisions follow a writing protocol familiar to the annotator (expert in law), there is no validation of the expert's annotations with an inter-agreement score. 
	
	Finally, it is important to note that it is a classical NER problem with four classes, nevertheless only the PER, LOC and DATE classes are useful for the de-identification problem. 

	
	\begin{table}[]
		\centering
		\makebox[\columnwidth][l]{
			\begin{adjustbox}{max width=1.0\columnwidth,  totalheight=3.4cm}
	
		\hspace*{-.3cm}
		\begin{tabular}{l|c|l|l|l|l}
			\multicolumn{2}{l|}{\textbf{Dataset}} & \textbf{Train} & \textbf{Dev} & \textbf{Test} & \textbf{Total} \\ \hline \hline
			\multicolumn{2}{l|}{\# of cases}  &     57          &   20           &     17       &     94   \\
			\multicolumn{2}{l|}{\# of sentences}  &    6,989          &  1,963           &   2,257        &    11,209     \\
			\multicolumn{2}{l|}{\# of tokens}     &    173,448           &    42,964        &     60,293       &   276,705                \\ \hline 
			\multirow{4}{*}{Ent}      & PER      &       1799         &     447         &        629     &    2875            \\
			& LOC      &             468   &         115     &    139           &       722         \\
			& PRO      &     750           &         215     &    243           &        1208        \\
			& DATE     &         57      &    9          &       18        &        84       
		\end{tabular}
		\end{adjustbox}
		}
		\caption{Description of the dataset of French court decisions with the associated entities}
		\label{tab:dataset} 
	\end{table}
	
%

	\subsection{Results}
	
	Regarding the metrics, we use the ratio of the true positives over the sums of the: true positives and false positives (precision), true positives and false negatives (recall), true positives and false positives and negatives (accuracy). The F1 is the weighted harmonic mean of the precision and the recall.


\begin{table*}[!ht]
	\centering
			\makebox[\columnwidth][l]{
				\hspace*{-3.cm}
		\begin{adjustbox}{max width=1.0\columnwidth,  totalheight=6.5cm}
			\centering
	\begin{tabular}{c|c|l|l|l|l}
		\textbf{Model}         &  \textbf{Context}          & \textbf{Rec} & \textbf{Prec} & \textbf{F1} & \textbf{Acc} \\ 	\hline \hline
		\textit{CRF-Baseline} \cite{Peng2014} & 0 & 79.06        & 92.77         & 85.37       & 74.47 \\
		MICA + \textit{CRF-Baseline}       & 8 &        80.31     &  94.28             &     86.74        &     76.58       \\
		MICA + \textit{CRF-Baseline}       & 128 &     \textbf{86.30}     &  \textbf{93.83}             &     \textbf{89.91}        &     \textbf{81.67}         \\
		MICA + \textit{CRF-Baseline}       & 512 &     87.39     &  92.30             &     89.78        &     81.45         \\		 \hline \hline 
		Vanilla \cite{Akbik2018} & 0 & 92.18        & 96.52         & 94.30       & 89.21        \\
		Vanilla + $\text{LM}_{fine tuned}$     & 0 &    93.62        & 96.11         & 94.85       & 90.20        \\ \hline \hline 
		MICA + $\text{LM}_{fine tuned}$        & 0 &     93.68     &  97.03             &     95.33        &     91.07         \\
		 MICA + $\text{LM}_{fine tuned}$   & 1 &   93.87      &   97.04             &   95.43          &      91.26       \\
		 MICA + $\text{LM}_{fine tuned}$   & 8 &   94.36      &   96.95             &   95.64          &      91.64       \\
		 MICA + $\text{LM}_{fine tuned}$   & 32 &   95.94      &   96.90             &   96.42          &      93.08       \\
		MICA + $\text{LM}_{fine tuned}$        & 128 &   \textbf{96.23}      &   \textbf{96.81}             &   \textbf{96.52}          &      \textbf{93.28}       \\
		MICA + $\text{LM}_{fine tuned}$  & 256-512 &     96.34   &      96.62         &        96.48     &     93.20        \\
		\hline 
	\end{tabular}
			\end{adjustbox}
}
	\caption{Results with the different models on the test dataset. The Context is in number of sentences.}
	\label{tab:results}
\end{table*}

	Table \ref{tab:results} reports the models' performances. First of all, our MICA enhanced CRF models obtain the best performances compared to their respective baselines. 
	
	Regarding the \textit{CRF-Baseline}, the results are still far from a state-of-the-art system like the Vanilla model of \cite{Akbik2018}. Nevertheless we can see that the MICA method is improving the results, even when applied on a \textit{CRF-Baseline}. We can note that for both the \textit{CRF-Baseline} and the BLSTM-CRF, when the context window is too wide, the precision of the system is dropping. For a window wider than 128, the gain in recall is not sufficient anymore to counter drop of precision in order to keep a high F1. 
	
	The Vanilla model of \cite{Akbik2018} obtains the poorest performances, but we can notice that using embeddings learned on legal domain rather than general domain helps significantly the system.
	
	Regarding our proposed models, we can notice difference of performances regarding the size of the context used to create the dictionary of entity candidates (see Equation \ref{eq:dico}). The best model is obtained with a context size of 128. 

	\paragraph*{High Recall} In the case of de-identification, we need our systems to reach a high recall in order to remove any sensible information. As a matter of fact, our best model allows a reduction of the relative recall error of 40,90\% compared with the fine-tuned Vanilla model.

	\subsection{Analysis}
	
	When analyzing our models, we witnessed several cases in which the systems we proposed improved the results over classical methods. To be more precise, we present in the subsequent few prediction  divergences between the model using a context of size 0 and our best model.
	\vspace*{-.2cm}
	\paragraph*{Typos} Obviously, our system allows to detect the entities with typos, as shown by the results highlighted in Table \ref{tab:results}. We noticed some relevant examples of missing spaces like the one below that were not detected by the model using no context: \\ \\ \vspace*{-.1cm}
	$\text{Whereas }\text{[MS.LAVERGNE]}_{PER}\text{ does not justify} \\ \text{her situation ...}$
	
	\paragraph*{Register} We noticed that when the register changes, the system can make mistakes. Especially when it happens that the entities to detect are children, they just use the first name to describe them, which is pretty uncommon for this kind of formal text. Our system can detect the name when it is presented for the first time in the text since there is the formality helping the system to detect the entity (Example 1), but struggle to detect the entity when it is in a long sentence without context (Example 2) :  
	$$ \text{\textbf{(1)} [J\'er\'emy]}_{PER}\text{ , born on February 19th, 1990.}$$ \vspace*{-.5cm} \\ 
	$\text{  \textbf{(2)} She states that }\text{[J\'er\'emy]}_{PER}\text{ and }\text{[L\'eo]}_{PER} \\ 
	\text{ have expressed the will to ....} $ \\
	
	One drawback of this method is that if the first model is predicting a false positive, it is likely that the new model will also predict that false positive. Nevertheless, the results do not show that behavior and the rate of false positive is stable. 
	Another drawback of our system is its inefficiency against the same words that were detected by the first models as different types of entities.
	
	\section{Conclusion and Future Works}
	\label{sec:ccl}
	
	In this paper, we introduced a new model for Named Entity Recognition. When tagging a sentence, It uses context elements in order to create a dictionary of entity candidates. This dictionary allows to compute a value corresponding to the potentiality of a word to be an entity of a certain type using the Damerau-Levenshtein distance and the longest common string distance to calculate a similarity coefficient. We tested our model on a dataset of French court decisions. Our results show a diminution of the relative recall error of more than 40\% compared to a fine-tuned state-of-the-art system while also slightly augmenting the precision. 
	
	
	We have in mind several improvements of our system, regarding the creation of the entities dictionary, the similarity function and the embeddings used. 
	
	A possible improvement of our system  to obtain a more accurate dictionary of entity candidates could be to 
	use the full document instead of a document-blind context window to create the dictionary of the entity candidates. We can see that a window size larger than 128 reduces the performance.
	
	Regarding the similarity function, it could be interesting to use the word embeddings generated by the character embeddings neural network with a cosine similarity. This would be an improvement over using only string-based similarities and take advantage of the robustness to noise of the character embeddings. 
	
	Recently, \cite{Edizel2019} proposed a new method to upgrade the Fastext embeddings in order to make them robust to misspelled words. It could be interesting to improve our system by replacing the classical Fastext embeddings into the vector of stacked embeddings with the ones of \cite{Edizel2019}. We leave this improvement for future work.  
	
	Finally, since our system is domain-agnostic and language-agnostic we strongly want to compare it on other classical open-domain NER datasets with different languages \cite{Tjong2003}.
	
	\section*{Acknowledgments}
	
	We would like to thank Pavel Soriano and C\'edric Malherbe for the insighful discussions and helpful comments. The authors of this work have been funded by the EIG program of the State in the context of the OpenJustice project.   
	
	
	\bibliographystyle{acl_natbib}
	\bibliography{OpenJustice}
	
\end{document}